\documentclass{article}

\usepackage{acra}
\usepackage{graphicx}
\usepackage{multirow}
\usepackage[table,xcdraw]{xcolor}
\usepackage{adjustbox}
\usepackage{amsmath}
\usepackage[hyphens]{url}
\usepackage{subcaption}
\usepackage{hyperref}
\usepackage{siunitx}
\usepackage[labelfont=bf]{caption}
\usepackage{tcolorbox}
\tcbuselibrary{breakable} 
\usepackage{fvextra}      

\usepackage{color, colortbl} 
\definecolor{Gray}{gray}{0.9} 

\title{Automated Marine Biofouling Assessment: Benchmarking Computer Vision and Multimodal LLMs on the Level of Fouling Scale}
\author{
  Brayden Hamilton$^{*}$, Tim Cashmore$^{*}$, Peter Driscoll, Trevor Gee, Henry Williams$^{**}$\\
  Centre for Automation and Robotic Engineering Science\\
  The University of Auckland, NZ\\
  \texttt{henry.williams@auckland.ac.nz$^{**}$} \\
}

\begin{document}

\maketitle

\let\thefootnote\relax\footnotetext{$^{*}$ These authors contributed equally to this work.}

\begin{abstract}    
    Marine biofouling on vessel hulls poses major ecological, economic, and biosecurity risks. Traditional survey methods rely on diver inspections, which are hazardous and limited in scalability. This work investigates automated classification of biofouling severity on the Level of Fouling (LoF) scale using both custom computer vision models and large multimodal language models (LLMs). Convolutional neural networks, transformer-based segmentation, and zero-shot LLMs were evaluated on an expert-labelled dataset from the New Zealand Ministry for Primary Industries. Computer vision models showed high accuracy at extreme LoF categories but struggled with intermediate levels due to dataset imbalance and image framing. LLMs, guided by structured prompts and retrieval, achieved competitive performance without training and provided interpretable outputs. The results demonstrate complementary strengths across approaches and suggest that hybrid methods integrating segmentation coverage with LLM reasoning offer a promising pathway toward scalable and interpretable biofouling assessment.    
\end{abstract}

\section{Introduction}
    New Zealand's marine ecosystems are at risk from the spread of invasive species, which can attach to the hulls of recreational and commercial vessels. Identifying and managing biofouling, the accumulation of aquatic organisms on submerged surfaces, is crucial for protecting biodiversity and safeguarding commercial fisheries \cite{biosecurityRisksBiofoulingMerchantVessels2004}. 

    Biofouling poses significant challenges to marine operations, vessel efficiency, and biosecurity. New Zealand regulatory frameworks, such as the Craft Risk Management Standard (CRMS), have increased the demand for scalable, objective, and accurate methods to assess vessel hulls to combat this challenge. Current biofouling survey methods are resource-intensive, relying on divers to visually inspect vessels, a practice that is both time-consuming and hazardous. The limited speed and scale of manual surveys mean that only a small number of vessels can be inspected. 

\begin{figure*}[!t]
    \centering
    \begin{subfigure}[t]{0.30\textwidth}
        \centering
        \includegraphics[width=\linewidth]{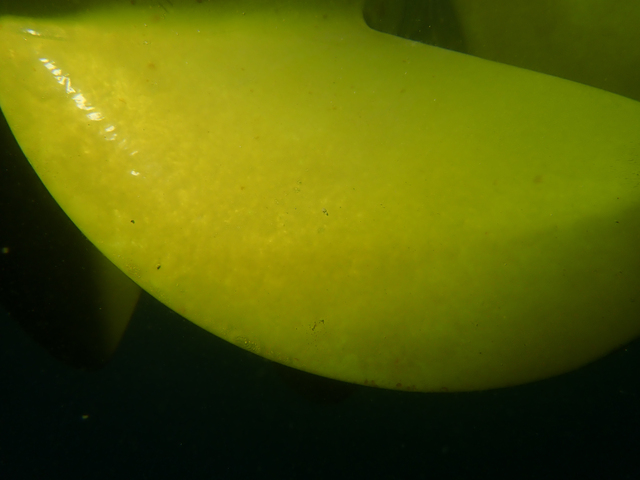}
        \caption{LoF 0: Clean surface, no visible fouling.}
    \end{subfigure}
    \hfill
    \begin{subfigure}[t]{0.30\textwidth}
        \centering
        \includegraphics[width=\linewidth]{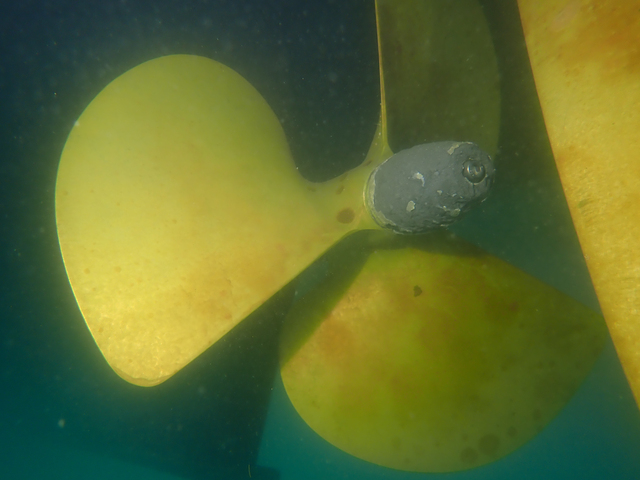}
        \caption{LoF 1: Slime layer only, no macrofouling.}
    \end{subfigure}
    \hfill
    \begin{subfigure}[t]{0.30\textwidth}
        \centering
        \includegraphics[width=\linewidth]{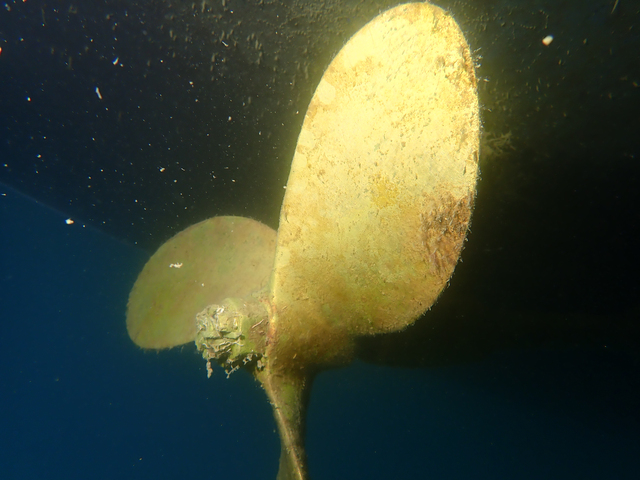}
        \caption{LoF 2: Sparse small organisms ($<5\%$ cover).}
    \end{subfigure}
    
    \vspace{0.3cm}
    
    \begin{subfigure}[t]{0.30\textwidth}
        \centering
        \includegraphics[width=\linewidth]{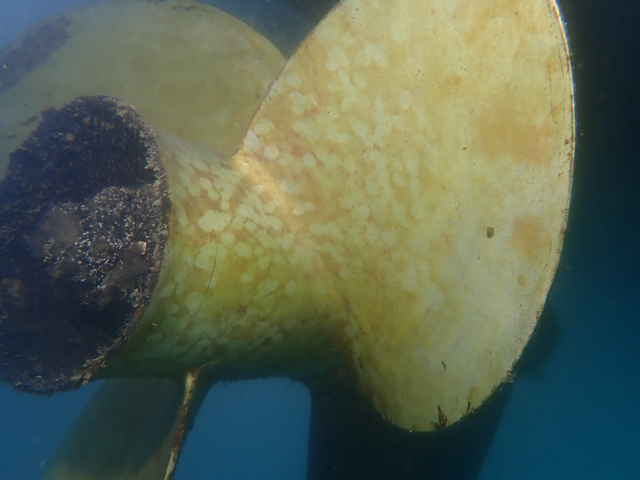}
        \caption{LoF 3: Moderate fouling patches ($5-16\%$ cover).}
    \end{subfigure}
    \hfill
    \begin{subfigure}[t]{0.30\textwidth}
        \centering
        \includegraphics[width=\linewidth]{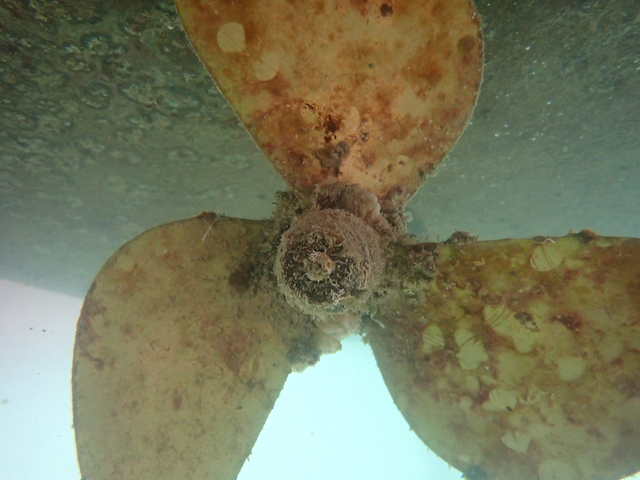}
        \caption{LoF 4: Extensive fouling ($16-40\%$ cover).}
    \end{subfigure}
    \hfill
    \begin{subfigure}[t]{0.30\textwidth}
        \centering
        \includegraphics[width=\linewidth]{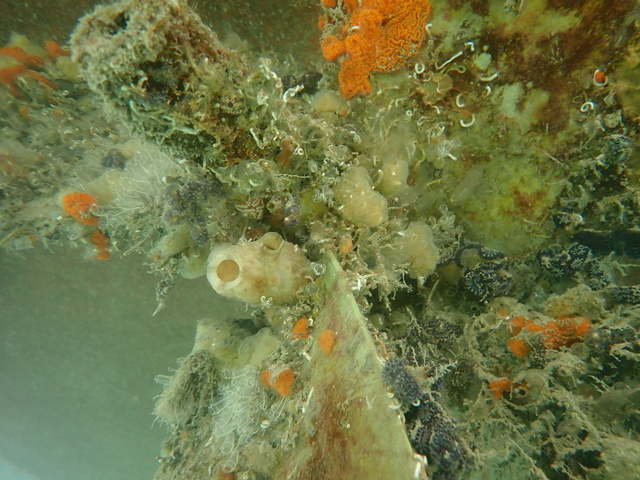}
        \caption{LoF 5: Heavy macrofouling ($>40\%$ cover).}
    \end{subfigure}
    
    \caption{Examples of propeller biofouling across the six Level of Fouling (LoF) categories (0–5). Each subfigure illustrates the progressive increase in fouling severity as defined by the LoF scale.}
    \label{fig:lof-scale}
    \end{figure*}

    This paper investigates automated classification methods for underwater biofouling detection on the Level of Fouling rank scale (LoF). The LoF rank scale, developed by \cite{Davidson_et_al_2019_Level_of_Fouling}, provides a practical visual framework for classifying biofouling severity and is applied to all vessels entering New Zealand. The levels themselves are described in detail in section \ref{sec:lof} and shown visually in Figure \ref{fig:lof-scale}. Utilising a combination of deep learning vision models and large multimodal language models (LLMs), the goal is to assess the effectiveness, generalisability, and practical trade-offs of these approaches in the context of marine biosecurity. 

    Drawing on the LoF guidelines as ground truth, two complementary experimental pipelines were investigated: (i) convolutional neural networks (ResNet-18/50) and transformer-based architectures (SegFormer) trained on expert-labelled datasets, and (ii) large language models accessed via the OpenRouter API, adapted through system prompt engineering and retrieval-augmented generation (RAG).

    The objective of this work is to develop a reproducible benchmarking framework for LoF classification and provide actionable insights for marine biofouling inspection systems. A systematic comparison is presented between convolutional neural networks, transformer architectures, and large multimodal language models, using a common expert-labelled dataset provided by the Ministry of Primary Industries. This study establishes a foundation for scalable, automated biofouling assessment and offers practical guidance for future deployment in marine biosecurity applications.

\section{Background and Related Work}
    Marine biofouling refers to the unwanted accumulation of aquatic organisms on submerged surfaces, including microbial films and macrofouling species, such as barnacles, Mediterranean fanworms, and Caulerpa. This process has significant ecological, operational, and economic consequences. On ships and other marine infrastructure, biofouling reduces hydrodynamic efficiency, increases drag, and increases fuel consumption, all resulting in increased greenhouse gas emissions \cite{EconomicImpactBiofoulingShips2011}. From a biosecurity perspective, ship hulls can act as a vessel for spreading invasive species, prompting increasingly strict regulations worldwide. In New Zealand, the Craft Risk Management Standard (CRMS) requires vessels to adhere to acceptable biofouling thresholds upon entry \cite{MPIVesselsCraftRiskManagement2023}, with the Level of Fouling (LoF) scale introduced by the Cawthron Institute as a standardised measure of fouling severity \cite{Davidson_et_al_2019_Level_of_Fouling}.

\subsection{Traditional Detection and Classification Methods}
    Current methods for detecting biofouling involve sending professional divers to manually inspect the hulls of vessels. This process is hazardous, labour-intensive, and does not scale well. Other traditional methods involve enhancing underwater images to aid in the identification of biofouling and specific species, such as contrast enhancement and colour correction \cite{underwaterImageEnhancement2021}, but this still requires a human assessment of the images. 
    
    Recent automated methods look at using deep learning techniques to identify marine fouling, with CNNs being popular among research \cite{underwaterObjectDetectionDatasets2024}. While these methods show promise, a common problem with using CNNs involves data scarcity. With a wide range of water, turbidity, and lighting conditions, it has been shown to be difficult to build models that are robust enough to handle all types of conditions. Furthermore, these issues, coupled with biases in dataset annotations, can heavily influence performance evaluations \cite{divingDeeperUnderwaterImageEnhancement2020}.

\subsection{Current Research Trends}
    Recent work in biofouling detection and classification shows three interlinked trends, which help to situate contributions and highlight gaps this paper aims to address.
    
    First, there is growing progress in fine-scale detection and classification, especially for slime/microfouling (e.g. diatoms) under varied environmental conditions. Models like DiatomNet \cite{DiatomNet2025} have been developed to classify diatom genera from large numbers of species, achieving good accuracy while remaining computationally efficient. Likewise, SC-DiatomNet \cite{Li2024_SC_DiatomNet} improves on earlier detection speed and accuracy for small fouling patches by using adaptive anchor boxes and spatial pooling to better handle diatoms of various sizes. These works show that, given enough labelled data and good imaging conditions, fine granularity is possible, but they also reinforce limitations when images are noisy, poorly illuminated, or when fouling features are subtle or partial.
    
    Second, there are broad surveys and roadmap studies that synthesise the state of ML methods for biofouling in infrastructure and marine energy applications, and which explicitly call out the major challenges around estimation, generalisation, and standardisation. The roadmap provided by \cite{Rashid2023_TST_Roadmap} provides a comprehensive overview of how biofouling has been detected and estimated for tidal stream turbine systems: it highlights that most detection models struggle with estimating the extent of fouling (coverage) rather than simply classifying presence/absence, and that environmental and imaging variability (lighting, turbidity, angle) continue to degrade performance. Another related study, by \cite{Rashid2024_TST_Classification}, extends this by applying ML approaches in field-relevant conditions, including transfer learning, data augmentation, and ensemble models to improve robustness.
    
    Third, two persistent challenges are repeatedly noted: (a) lack of standardised benchmarks, including agreed severity/extent definitions and consistent inter-expert annotation agreement; and (b) limited generalisation across variable imaging settings. For example, the roadmap work \cite{Rashid2023_TST_Roadmap} emphasises that many studies report high accuracy in controlled lab or short-term field studies but perform poorly in long-term, low visibility, or adverse conditions. Likewise, the diatom classification work often depends on high-resolution, clean images, which do not always represent typical underwater inspection scenarios.

\subsection{Emergence of LLMs and Multimodal Models}
    Large language models (LLMs) have shown remarkable capabilities in processing unstructured data such as reports, logs, and ecological metadata. With the advent of multimodal LLMs, such as GPT-4V, it is now possible to integrate visual and textual reasoning, offering new possibilities for marine and ecological applications. \cite{zheng2024exploringboundarygpt4vmarine} conducted a case study on GPT-4V for marine analysis, demonstrating its ability to recognise broad patterns and interpret marine imagery, while also highlighting limitations in fine-grained species identification and boundary cases. Similarly, \cite{yoshida2024jediqabenchmarkdeepsea} introduced the J-EDI QA benchmark for evaluating multimodal LLMs on deep-sea organism images, finding that models performed below expert level, particularly in species-specific recognition. 
    
    Recent efforts to domain-specialise these models underscore their potential. \cite{bi2024oceangptlargelanguagemodel} developed OceanGPT, a large language model tailored for ocean sciences, and benchmarked it on diverse oceanographic tasks through the OCEANBENCH framework. Their results illustrate that domain-specific pretraining improves reasoning and robustness compared to general-purpose LLMs. In related environmental contexts, multimodal models have been benchmarked for hydrological tasks such as flood severity estimation and water quality monitoring \cite{Kadiyala2024}, showing strong promise for integrating visual cues with contextual metadata. Collectively, these works indicate that multimodal LLMs are promising but still face challenges in domain generalisation, calibration, and fine-grained recognition in marine settings.

\subsection{Opportunities for LLMs in Biofouling}
    Despite promising progress, the use of LLMs in biofouling research remains nascent. Prior work on automated biofouling assessment has primarily focused on conventional deep learning approaches. \cite{Mannix2021} developed a large-scale image-based pipeline for severity classification using expert consensus as ground truth, demonstrating feasibility but limited generalisability under variable conditions. More recent approaches, such as interpretable prototype-based methods for estimating coverage, offer improvements in explainability and alignment with human judgement \cite{mannix2025interpretableapproachautomatingassessment}. 
    
    LLMs open new avenues for integrating non-image data, such as vessel metadata, inspection site, and expert notes, with visual classification. In marine science, \cite{Schoening2022} emphasised the importance of standardising image metadata and annotation formats to enable reproducible ecological analysis. Semi-supervised and self-supervised learning also present opportunities to leverage large volumes of unlabeled underwater imagery, as demonstrated by \cite{Ding2025}, who applied multimodal contrastive learning to enhance underwater images in variable conditions. Relatedly, \cite{Miao2025} showed how multimodal LLMs can generalise to novel ecological data modalities such as bioacoustics, reinforcing the potential of LLMs to unify text, metadata, and visual data streams. These opportunities suggest that LLMs could play a central role in future biofouling detection systems by combining image analysis with contextual information to standardise severity classification and support automated reporting workflows.

\subsection{Key Challenges}
    Both conventional methods and LLM-based approaches to image classification face several unique challenges. Deep neural networks and LLMs are often very computationally expensive, making integration into devices with tight energy requirements a key limitation of these technologies \cite{RealTimeClassifcation2020}. Challenges with conventional approaches to underwater image classification lie with the fundamental properties of the underwater environment, where a wide variety of lighting and water turbidity can result in significant differences in the appearance of an object \cite{deepLearningUnderwaterMarineObjectDetection2023}. While there are techniques to mitigate these issues, they still pose a major challenge for deep neural networks and image classification tasks in the underwater environment \cite{deepLearningUnderwaterMarineObjectDetection2017}. Furthermore, biases in image annotations can lead to skewed results in these models \cite{divingDeeperUnderwaterImageEnhancement2020}.

    In contrast, LLM-based approaches to image classification often face issues regarding fine-tuning prompts and hallucinations \cite{llmsAsVisualExplainers2023}, in addition to challenges associated with visually challenging datasets, like in underwater imagery, as mentioned above. Finally, LLMs and other AI systems face issues with interpretability and trust \cite{enhancingTrustLLMs2024}. For regulatory and operational acceptance, AI systems must provide interpretable results that can be trusted by human operators and professionals.

\section{Methods}\label{sec:methods}
    The purpose of this methodology is to develop and benchmark automated approaches for classifying marine biofouling severity on vessel hulls. Utilising the Level of Fouling (LoF) rank scale \cite{Davidson_et_al_2019_Level_of_Fouling}, expert-labelled imagery was combined with both traditional computer vision pipelines and large multimodal language models (LLMs). This approach proceeds in four stages: (i) establishing LoF as the classification framework, (ii) preparing and enriching the dataset with segmentation annotations, (iii) designing and training conventional deep learning models, and (iv) evaluating zero-shot multimodal LLMs through structured prompting and retrieval-augmented generation. This structure allows direct comparisons between model families under identical conditions, assesses their generalisation to variable underwater imagery, and explores their potential for operational deployment in marine biosecurity.

\subsection{Level of Fouling}\label{sec:lof}
\begin{figure}[ht]
        \centering
        \includegraphics[width=0.60\linewidth]{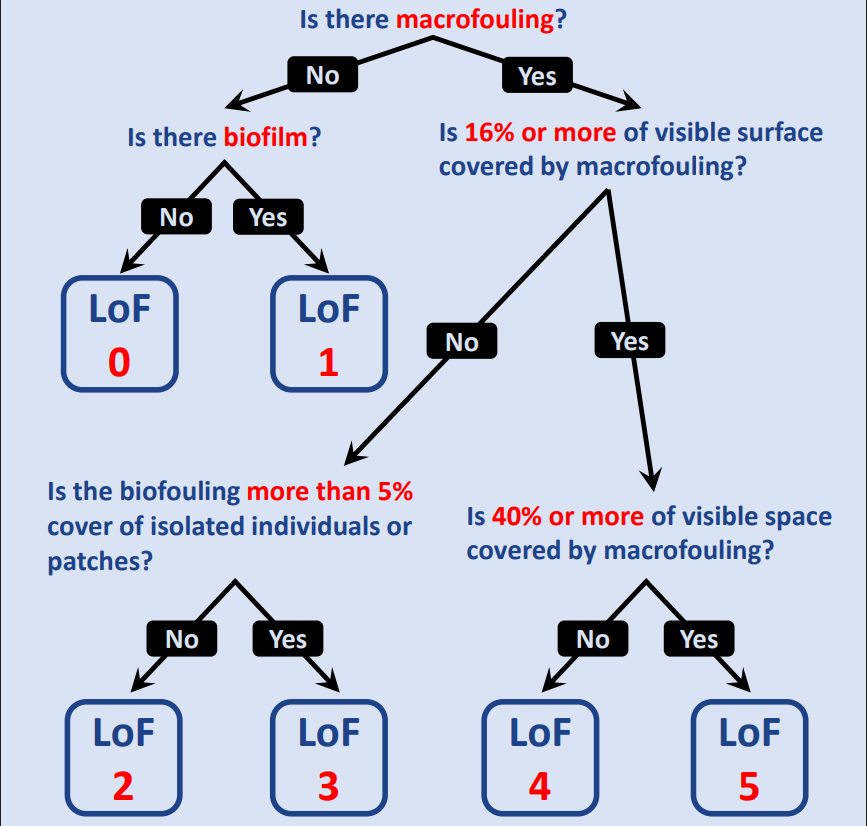}
        \caption{Decision tree for determining the level of fouling of a ship given the observations of the hull.}
        \label{fig:lof_flow}
    \end{figure}

    The Level of Fouling (LoF) Rank Scale is a standardised framework for categorising the extent of biofouling on vessel hulls and other submerged surfaces. Developed by \cite{Davidson_et_al_2019_Level_of_Fouling} for practical application in marine biosecurity, the scale ranges from 0 (no visible fouling) to 5 (heavy macrofouling), providing a rapid and consistent means of assessing fouling pressure without requiring detailed species identification. The specific description of each level is presented in Figure \ref{fig:lof-scale} with the decision tree for level classification shown in Figure \ref{fig:lof_flow}. These levels are the classification labels that these methods intend to classify for each image. Raw image classification will learn the LoF scale directly; semantic classification methods provide the data necessary to work through the existing expert decision tree shown in Figure \ref{fig:lof_flow}. LLM-based methods use the expert decision tree directly to classify using its own estimates of biofouling coverage.


\subsection{Dataset and Annotations}
    The dataset used in this study was provided by New Zealand's Ministry for Primary Industries (MPI) and consists of underwater photographs of recreational and commercial vessel hulls captured during routine surveillance of vessels. Each image was annotated with an expert-assigned Level of Fouling (LoF) ranking (0-5), serving as the ground truth for classification. In total, 762 images were provided, the LoF class distribution of which is provided in Table \ref{tab:lof_dist}. 

    \begin{table}[ht]
    \centering
        \begin{tabular}{c c}
        \hline
        \textbf{LoF} & \textbf{Number of Images} \\ \hline
        0 & 7 \\
        1 & 263 \\
        2 & 70 \\
        3 & 113 \\
        4 & 126 \\
        5 & 183 \\ \hline
        \textbf{Total} & 762 \\ \hline
        \end{tabular}
    \caption{Distribution of images across Level of Fouling (LoF) classes.}
    \label{tab:lof_dist}
    \end{table}

\subsection{Raw Image Classification}
    A supervised deep learning pipeline for LoF classification was developed, centred on convolutional neural networks and transformer-based segmentation models. Two CNN backbones (ResNet-18 and ResNet-50) were evaluated for their trade-off between speed, generalisation, and fine-grained detail capture. ResNet-18 and ResNet-50 have demonstrated robustness in underwater environments \cite{underwaterResNet2022}. ResNet-18's architecture helps extract features in noisy underwater environments, while ResNet-50 provides deeper feature extraction, at the cost of being more computationally expensive \cite{ResNet18vs50-2025}. These standard image classifiers serve as the baseline, offering a vision-only benchmark without semantic reasoning or domain knowledge, against which VLLM-based and semantic rule–augmented classifiers are compared. 

\subsection{Semantic Classifier}
\begin{figure}[ht]
        \centering
        \includegraphics[width=1\linewidth]{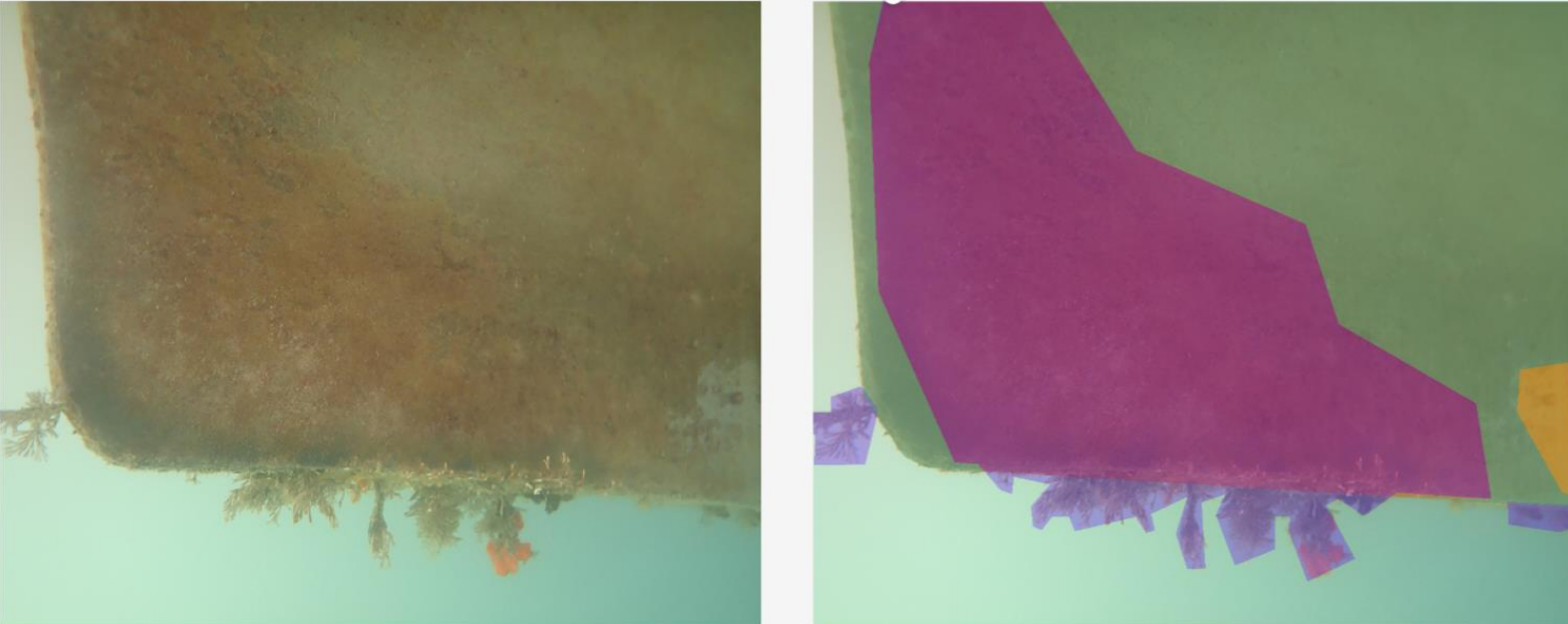}
        \caption{Example of an image with class labels, where Macrofouling is purple, Slime is green, and Clean is yellow.}
        \label{fig:labelling-example}
    \end{figure}

    In addition, SegFormer, a transformer-based semantic segmentation model that has been shown to outperform other transformer models in the underwater environment \cite{segformer2025}, was applied to learn pixel-level masks of the key features used for LoF classification. Pixel-level segmentation masks were introduced with four mutually exclusive classes defined: Water, Clean, Slime, and Macrofouling, as shown in Figure \ref{fig:labelling-example}. These were manually applied using annotation tools and enabled two downstream benefits: (i) improved interpretability of model predictions by explicitly separating areas of the hull from background water, and (ii) the ability to calculate precise coverage estimates for slime and macrofouling, which can be mapped directly to LoF thresholds.  

     The model was used to determine the coverage of macrofouling, slime, and clean patches, which were then used to estimate the area of the boat hull and, consequently, the percentage coverage of macrofouling and slime. The predicted biofouling coverage percentages could then be directly mapped onto the LoF rules outlined in Figure \ref{fig:lof-scale}, alongside highlighting areas affected by biofouling.

\subsection{HSV and Edge Extraction}
    In addition to testing raw image inputs, pre-processing techniques designed to highlight colour and texture features that are particularly important in underwater imagery were explored. The rationale was that handcrafted computer vision models often struggle to separate slime from macrofouling under conditions of poor lighting, turbidity, or colour distortion, and enhancing input features could make these features more salient during training. This pre-processing stage was tested for both the standard image classifier and the segmentation approaches to determine if they enhanced the performance over using the raw image.

    The first technique introduced was HSV colour space extraction. By converting images from RGB to HSV, it was possible to isolate hue and saturation information, which more closely represents how fouling organisms differ visually from clean hull surfaces under variable underwater lighting. These HSV channels were provided to the network alongside the original image, giving models an additional feature space to distinguish subtle differences in slime layers, macrofouling patches, and clean surfaces.

    The second technique was edge detection, which was applied as an additional input channel. This focused on enhancing structural and textural information by highlighting object boundaries and surface discontinuities that are not as easily captured in raw pixel intensity values. Fouling organisms such as barnacles and tubeworms typically produce distinctive edges against smooth hull surfaces, and edge-enhanced inputs helped all models to exploit these cues.


\subsection{Model Training}
    Both the classification models and segmentation models were trained under the same hyperparameters, as outlined in Table \ref{tab:training}. These hyperparameter values were identified through iterative testing of the models. Similarly, the dataset was split into 80\%-20\% training and testing sets, respectively. The dataset was split only once to ensure that all models were trained and tested on the same information, thereby maintaining consistent results.

    \begin{table}[ht]
    \centering
        \begin{tabular}{l c}
        \hline
        \textbf{Hyperparameter} & \textbf{Value} \\ \hline
        Epochs & 60 \\
        Batch size & 32 \\ 
        Learning rate & 0.001\\ \hline
        \end{tabular}
        \caption{Training hyperparameters used for all supervised models.}
        \label{tab:training}
    \end{table}

\subsection{LLM-Based Pipeline}
    A pipeline based on vision-capable large language models (LLMs), accessed programmatically through the OpenRouter API, was developed. This approach reframes LoF classification as a multimodal reasoning task, enabling the use of pretrained models without fine-tuning. Each underwater image was converted into base64-encode format, a model-agnostic method that ensures consistent ingestion across different providers.
        
    Structured system prompts were designed that: (i) defined the LoF scale with precise thresholds, (ii) instructed the model to provide a discrete integer output, and (iii) framed within a marine biosecurity context (``You are a marine biosecurity expert''). Iterative refinements explored conservative vs. balanced prompt formulations to mitigate systematic over- or under-classification at specific LoF levels. 
    
    Furthermore, Retrieval-Augmented Generation (RAG) was used to strengthen alignment with domain standards; experimentation was conducted by injecting excerpts from the official LoF guideline into prompts. This included scale definitions, examples of slime vs. macrofouling, and boundary criteria for each LoF category. The baseline prompt and the final system prompt are provided in Appendix \ref{app:best-prompt}. 
        
    
\section{Results}
    To ensure fair comparison, all pipelines were evaluated on the same dataset. The supervised models were trained on labelled subsets with identical train-test splits, while LLMs operated in a zero-shot regime. In addition to accuracy, robustness to variable image quality (turbidity, lighting, resolution) and interpretability of outputs were carefully monitored dtouring evaluation. 



\subsection{Raw Image Classification}
\begin{figure*}[!t]
    \centering
    \begin{subfigure}[t]{1\textwidth}
        \centering
        \includegraphics[width=1.0\linewidth]{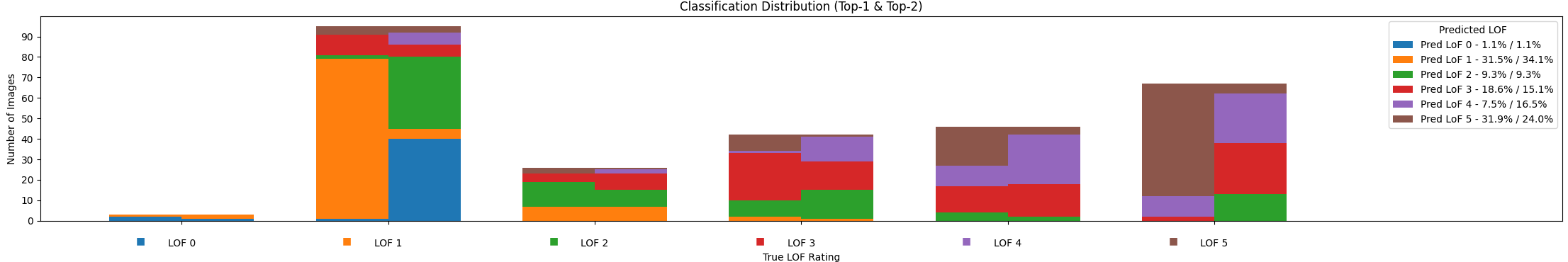}
        \caption{Results for ResNet-18 without HSV colouring.}
        \label{fig:resnet18-no-hsv}
    \end{subfigure}
    \vspace{0.3cm}
    \begin{subfigure}[t]{1\textwidth}
        \centering
        \includegraphics[width=1.0\linewidth]{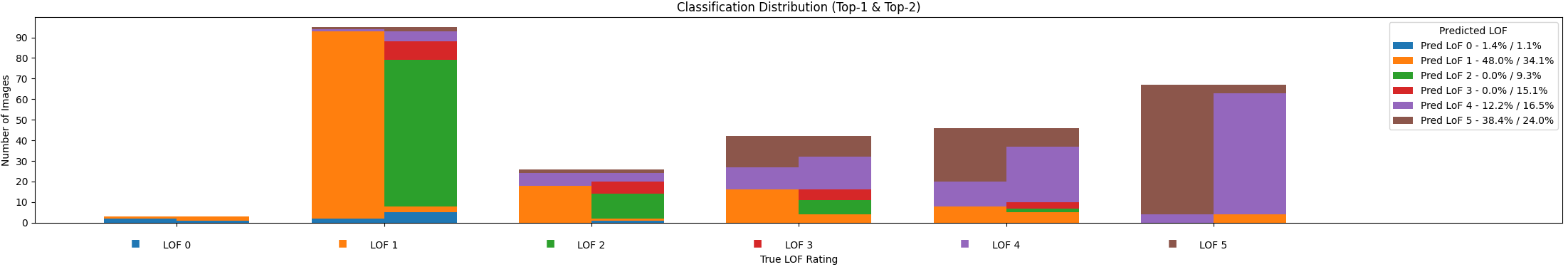}
        \caption{Results for ResNet-50 without HSV colouring.}
        \label{fig:resnet50-no-hsv}
    \end{subfigure}
    
    \caption{Each column represents the number of images in each respective class, where the colour of the column represents the predicted LoF rating. The left side represents the most confident prediction, while the right side represents the second most confident prediction.}
    \label{fig:resnet-model-graphs}
    \end{figure*}

    Initial tests of these models showed a high accuracy in LoF 1 and LoF 5 predictions, whereas the models struggled in differentiating between LoF 2, 3, and 4. LoF 0 images were often predicted to be LoF 1. Due to the unexplainability inherent to this approach, it is difficult to diagnose the specific cause of the inaccuracies in these models. Results shown in figure \ref{fig:resnet-model-graphs} show that ResNet-50 excels at LoF 1 and 5, while ResNet-18 excels at LoF 2 and 3. While class imbalance was likely the cause of above-average LoF 1 and below-average LoF 0 predictions, the lack of explainability of this method makes it unlikely to be accepted by industry. To solve this issue, LLM solutions and segmentation models were pursued concurrently. 

\subsection{Semantic Segmentation}
\begin{figure}[htb]
        \centering
        \includegraphics[width=\linewidth]{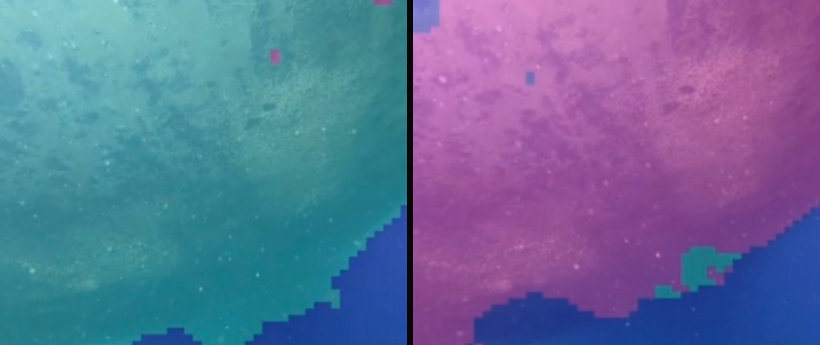}
        \caption{Different output predictions across two consecutive frames of video inference showing the instability in the micro- and macro-fouling predictions.}
        \label{fig:frames}
    \end{figure}

     The semantic segmentation model struggled to differentiate between macrofouling and slime within the boat area, particularly when a combination of these areas was present within the frame. Video inference revealed the model frequently switching between macrofouling and slime across consecutive and near-identical frames. An example of this is shown in Figure \ref{fig:frames}, where the classification flips between two consecutive frames. 

    Predicted percentage coverage of macrofouling and slime across the testing dataset revealed that the model often favoured either 100\% slime or 100\% macrofouling, with very few cases being a mix of both fouling types. A weighted average was applied to the video inference output based on the prediction confidence for each class. While this improved the consistency of the edges between macrofouling and slime, the model outputs still struggled to differentiate between the two classes. This performance makes it unsuitable for further LoF classification, leading to over-classifications for LoF 1 and 5. In an attempt to improve the contrast between macrofouling and slime, HSV and edge extraction were implemented.

    \begin{figure}[htb]
        \centering
        \includegraphics[width=0.8\linewidth]{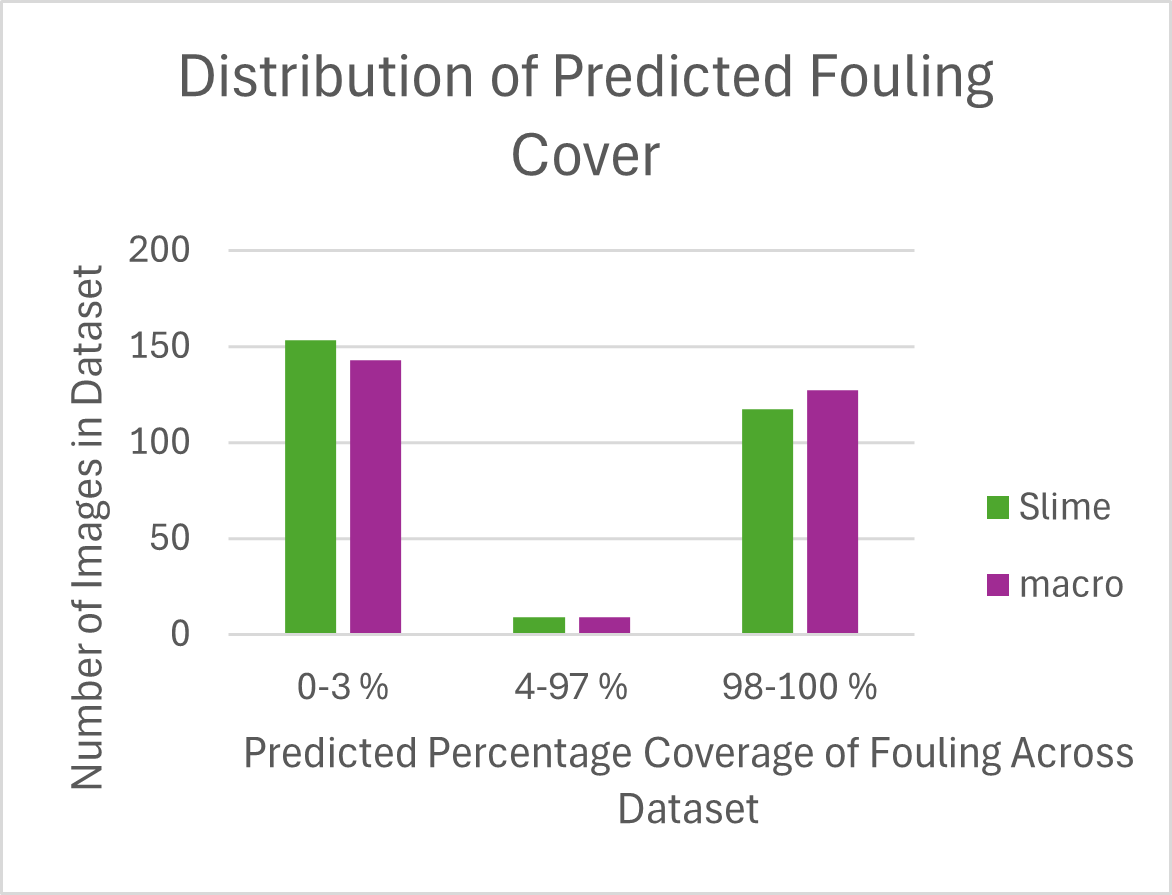}
        \caption{Distribution of Predicted Fouling Cover across the Dataset for the segmentation models.}
        \label{fig:dist_graph}
    \end{figure}

\subsection{HSV and Edge Extraction}
    The preprocessing steps were utilised to train classification models to understand if both HSV and edge extraction had an impact on alternative models to segmentation models. Improvements were observed when using these preprocessing steps when compared to both baseline classification models with the same hyperparameters. Clear improvements were in the ability to distinguish between LoF 2, 3, and 4, which was a major limitation of raw image classification. Overall accuracy on the test set of images increased minimally from 60.22\% to 62.72\%. The ability of the model to differentiate better on LoF boundaries shows that utilising HSV and edge extraction could be beneficial to differentiating between micro- and macro-fouling.

    \begin{figure*}[!t]
    \centering
    \begin{subfigure}[t]{1\textwidth}
        \centering
        \includegraphics[width=1.0\linewidth]{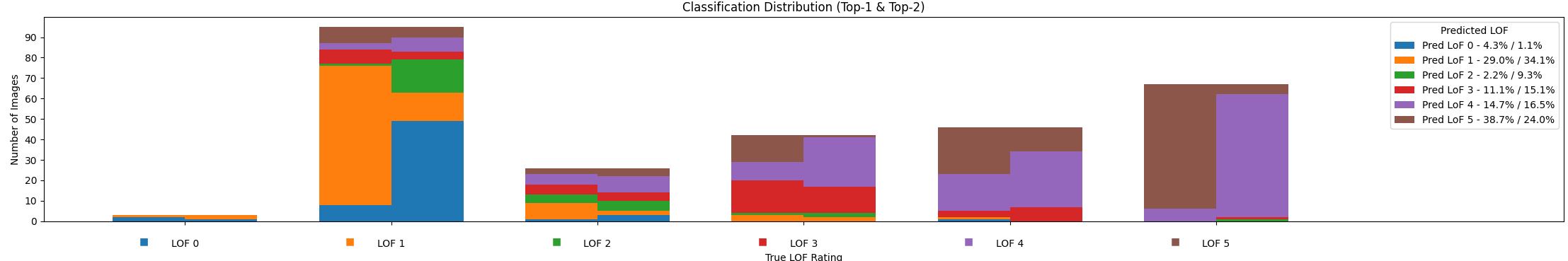}
        \caption{Results for ResNet-18 with HSV colouring.}
        \label{fig:resnet18-no-hsv}
    \end{subfigure}
    \vspace{0.3cm}
    \begin{subfigure}[t]{1\textwidth}
        \centering
        \includegraphics[width=1.0\linewidth]{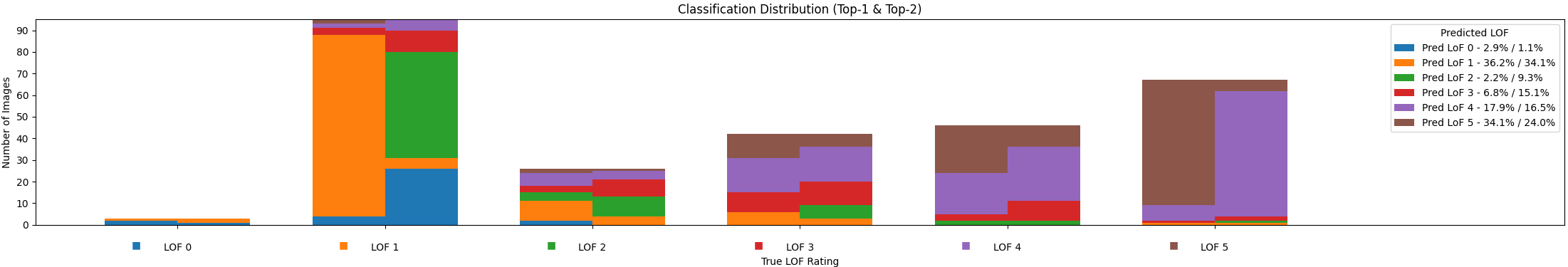}
        \caption{Results for ResNet-50 with HSV colouring.}
        \label{fig:resnet50-no-hsv}
    \end{subfigure}
    
    \caption{Each column represents the number of images in each respective class, where the colour of the column represents the predicted LoF rating. The left side represents the most confident prediction, while the right side represents the second most confident prediction.}
    \label{fig:resnet-model-graphs-hsv}
    \end{figure*}

    Across segmentation models implementing HSV colour and edge extraction to the preprocessing steps saw very little improvement in both overall accuracy and the model's ability to separate between slime and macrofouling effectively. When conducting video inference results, it was clear that the segmentation models still favoured 100\% slime or 100\% macrofouling. Overall, the results highlight the primary challenge in differentiating micro- and macro-fouling.
    
\subsection{LLM Evaluation Stages}
    To systematically evaluate Large multimodal Language Models (LLMs) for the task of Level of Fouling (LoF) classification, a series of staged experiments was designed. Each stage reflected increasing levels of control over the prompt design and input formatting, enabling identification of the key factors that influenced performance.
    
    

    Using only a simpler user-level prompt, classification rates were extremely low (5.1\% of images returned a label) and accuracy was poor, indicating that additional system-level conditioning was essential. 

    A detailed system prompt was introduced, including explicit LoF definitions, task framing (``You are a marine biosecurity expert''), and strict numerical output requirements. This substantially improved performance: 
    
    94.8\% of images received a classification, with an overall of 51.1\%. Accuracy was highest at the extremes (LoF 0, LoF 1, and LoF 5), but boundary cases (LoF 2-3, LoF 3-4) remained problematic, often over-classified into higher categories.
    
    To address systematic over-classification, prompts emphasising stricter thresholds for higher LoF categories were defined. This improved accuracy for low-level fouling (LoF 1 accuracy rose to 75.5\%), but reduced accuracy for LoF 4-5. Overall accuracy dropped to 42.7\%, demonstrating a trade-off between conservative calibration and balanced performance across all categories.

    \begin{figure*}[ht]
    \centering
    \begin{subfigure}[t]{0.45\textwidth}
        \centering
        \includegraphics[width=\linewidth]{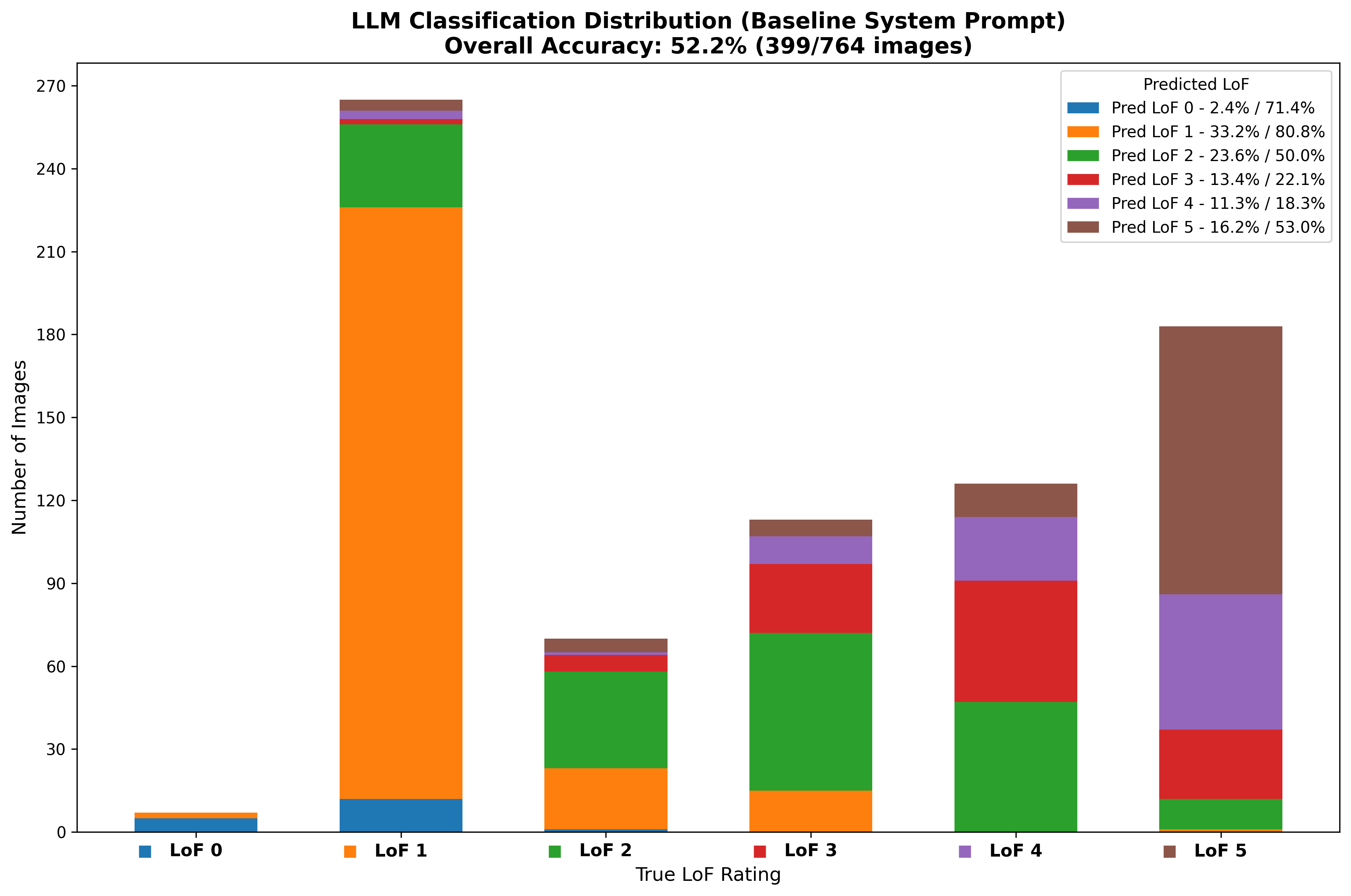}
        \caption{Test results showing classification accuracy with baseline system prompt}
        \label{fig:baseline}
    \end{subfigure}    
    \hfill
    \begin{subfigure}[t]{0.45\textwidth}
        \centering
        \includegraphics[width=\linewidth]{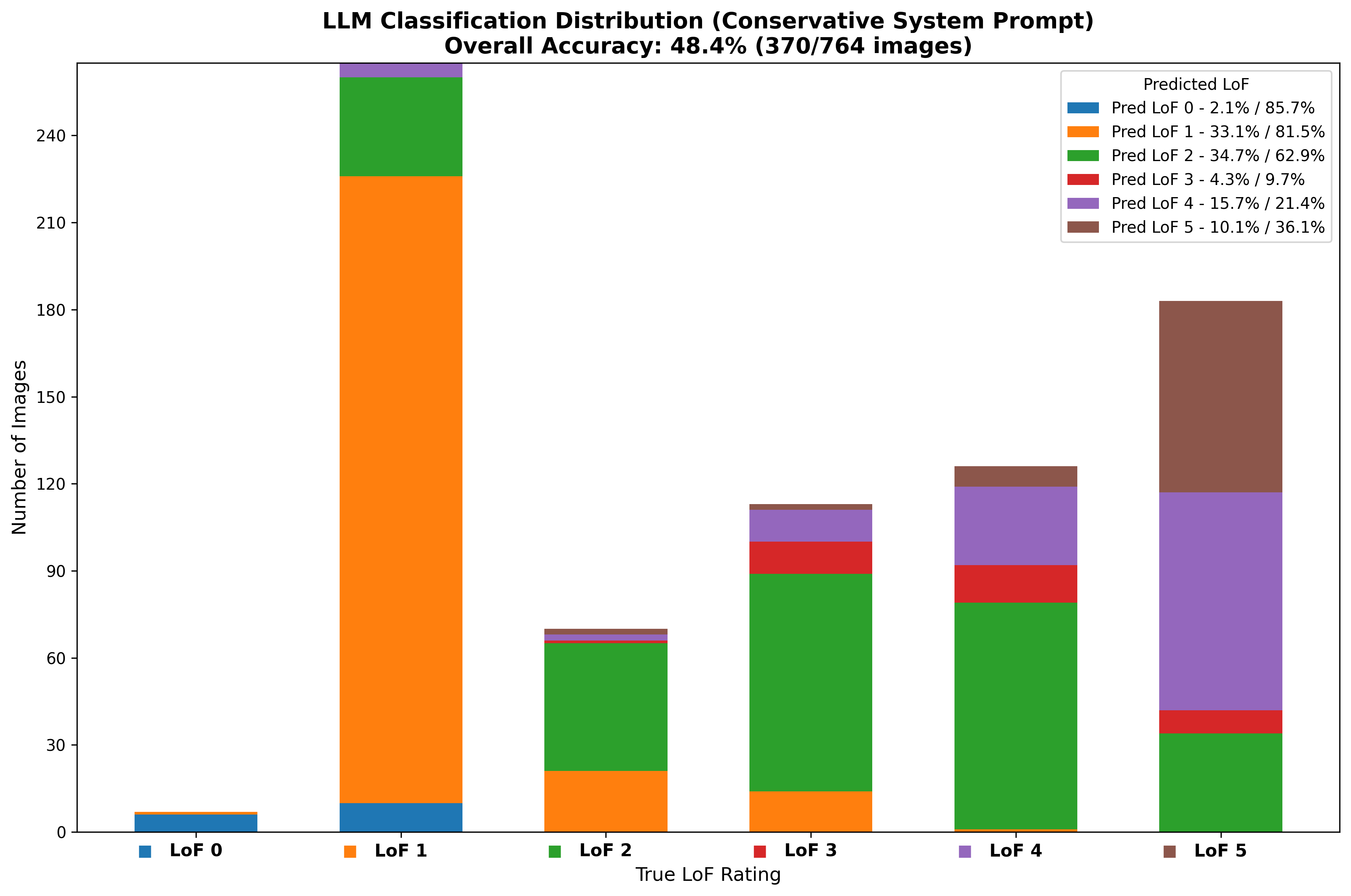}
        \caption{Conservative prompting strategy results — under-classification of higher LoF categories}
        \label{fig:conservative}
    \end{subfigure}
    \caption{Comparison of prompting strategies and baseline performance for LoF classification.}
    \label{fig:results-comparison}
\end{figure*}

\section{Discussion}
    It was observed that classification models achieved the highest accuracy in mapping images to the LoF scale compared to other methods. Although the segmentation and LLM approaches presented interpretability in their results, which may be more favourable to those without expertise in the field.
    For supervised computer vision models, CNN classifiers such as ResNet-18 and ResNet-50 achieved strong performance on extreme LoF categories, particularly LoF 1 and LoF 5, but struggled with intermediate levels (LoF 2–4). These misclassifications were exacerbated by class imbalance in the dataset and the black-box nature of classification outputs, which limited interpretability. Incorporating segmentation masks through transformer architectures (SegFormer) provided greater transparency by producing explainable overlays of micro- and macro-fouling percentage coverage estimates. However, the segmentation models often defaulted to 100\% coverage predictions for slime or macrofouling, failing to capture mixed states typical of LoF 2–4. While HSV colour and edge extraction provided modest improvements in classification models by making intermediate categories more separable, their overall effect was limited when applied to segmentation models, suggesting that input quality and dataset scope remain key bottlenecks. 

    The LLM-based pipeline revealed a different set of behaviours. Zero-shot classification via API calls initially failed to return useful results, underscoring the importance of structured system prompts. With explicit LoF definitions and role framing, LLMs demonstrated contextual awareness, producing accurate classifications for clean and heavily fouled images. Unlike CNNs, LLMs also provided explainable outputs in the form of textual reasoning, including descriptions of visible coverage and, in some cases, recognition of invasive species. This interpretability is valuable for operational use, as it gives inspectors insight into how the model reached a decision. Yet, similar to CNNs, LLMs struggled with transitional LoF boundaries, often over- or under-classifying intermediate cases depending on prompt calibration. Conservative prompt formulations improved LoF 1 accuracy but introduced underestimation at higher fouling levels, revealing the sensitivity of LLM predictions to small wording changes. Experiments with retrieval-augmented generation (RAG) suggested potential for grounding LLM outputs in domain guidelines, but the models did not consistently attend to injected context in long prompts. 

    Taken together, these findings illustrate that current methods are well-suited for identifying the extremes of the LoF spectrum but face persistent challenges in resolving ambiguous or borderline cases. They also underscore broader issues of dataset quality: class imbalance, image resolution, and hull coverage significantly affect both hand-crafted computer vision models and LLMs. At the same time, the interpretability of segmentation overlays and the textual explainability of LLMs both point to viable paths forward. Hybrid approaches, for example, combining classification results with pixel-level segmentation and LLM reasoning over contextual features, may help bridge the observed gaps.

    It is also worth noting that a limitation with this work is that the LoF rating is decided by the coverage of fouling across the entire boat hull, while the images in the dataset were close-up frames. Mapping of the ship hulls would be required for future work to properly estimate the coverage of fouling across the hull of the ship, and not just the immediate snapshot. 
    
\section{Conclusions and Future Work}
    This work compared automated biofouling classification methods, evaluating supervised computer vision pipelines and large multimodal language models (LLMs) against the Level of Fouling (LoF) framework. Key contributions include a systematic benchmark of CNNs, transformer-based segmentation models, and LLMs on a shared, expert-labelled dataset.

    Results showed that CNNs were efficient and robust for clear-cut cases but lacked interpretability and struggled with intermediate LoF levels. Transformer-based models produced explainable coverage maps but were limited by dataset framing. LLMs demonstrated flexibility and contextual reasoning, accurately identifying extreme LoF levels without training and offering interpretable textual outputs, though they were highly sensitive to prompt design and context length. None of the approaches fully addressed boundary-level classification, highlighting the need for improved datasets and hybrid approaches.
    
    Future work should focus on three directions: (1) expanding and balancing datasets at intermediate LoF levels, (2) embedding domain-specific context into LLM workflows via structured RAG pipelines or prompt templates, and (3) developing hybrid systems that combine segmentation-based quantitative analysis with LLM-driven reasoning. Such methods could enhance interpretability, scalability, and compliance with frameworks like CRMS, enabling faster and more reliable vessel inspections for marine biosecurity.

\section*{Acknowledgements}
    This project is sponsored by K\=otare Group Limited\footnote{\url{https://www.kotarenz.com/}}. Their support has been instrumental in enabling the research. Thanks to Biosecurity New Zealand, the Ministry for Primary Industries, the Auckland Council, the Bay of Plenty Regional Council, and Cawthron Institute for providing the imagery data used in this project.

\bibliography{publications}
\bibliographystyle{named}

\clearpage 

\clearpage
\clearpage
\onecolumn

\appendix

\section{Example Prompts}\label{app:prompts}

\begin{tcolorbox}[colback=gray!5,colframe=black!50,
                  title=Final System Prompt (Best Performing),
                  breakable,width=\textwidth]
\label{app:best-prompt}
\begin{Verbatim}[breaklines=true]
System Prompt:

You are a marine biofouling assessment expert with expertise in the standardized Level of Fouling (LoF) classification system. Your primary goal is ACCURATE LoF classification across all ranking levels (0-5).

LEVEL OF FOULING (LoF) ASSESSMENT FRAMEWORK:
Follow the official LoF Rank Scale methodology exactly:

LoF 0: No slime, no macrofouling - completely clean surface
LoF 1: Slime layer present (any amount), no macrofouling visible
LoF 2: 1-5% surface covered by macrofouling, patchy distribution or isolated individuals
LoF 3: 6-15% surface covered by considerable macrofouling
LoF 4: 16-40% surface covered by extensive fouling (more than half surface still clean)
LoF 5: 41-100% surface covered by very heavy macrofouling

STANDARDIZED ASSESSMENT PROTOCOL:
1. Examine the visible submerged surface systematically
2. Determine fouling type: absent, slime-only, or macrofouling present
3. If macrofouling present, estimate percent cover of the visible surface
4. Apply the appropriate LoF rank based on percent cover thresholds

KEY DEFINITIONS:
- Slime layer (biofilm): Microscopic organisms forming thin films or filaments
- Macrofouling: Visible organisms such as barnacles, algae, sponges, bivalves, sea squirts
- Percent cover: Proportion of visible surface occupied by fouling organisms
- Visible surface: The directly observable area in the image

DECISION TREE APPLICATION:
1. Is there any slime visible? 
   - If NO slime → LoF 0
   - If slime ONLY (no macrofouling) → LoF 1
2. If macrofouling is present, estimate percent cover:
   - 1-5% → LoF 2
   - 6-15% → LoF 3  
   - 16-40% → LoF 4
   - 41-100% → LoF 5

ASSESSMENT GUIDELINES:
- Focus on actual biological organisms, not surface discoloration or shadows
- Consider the full visible surface area when estimating coverage
- Distinguish between organic growth and inorganic deposits/staining
- Account for image perspective but base estimates on what is clearly visible
- Both hull areas and niche areas (rudders, propellers, complex structures) follow the same LoF criteria

NEW ZEALAND PRIORITY INVASIVE SPECIES:
When identifying species, focus on these high-risk organisms:
- Didemnum vexillum (Sea Vomit): Tan/beige/cream lumpy encrusting patches
- Sabella spallanzanii (Mediterranean Fanworm): Fan-like feeding crowns from tubes
- Undaria pinnatifida (Asian Kelp): Brown seaweed with broad fronds
- Styela clava (Asian Sea Squirt): Club-shaped solitary tunicates
- Ciona intestinalis (Sea Vase): Translucent cylindrical tunicates

RESPONSE REQUIREMENTS:
1. LoF rating (0-5) with clear justification based on percent cover
2. Coverage percentage estimate of visible surface
3. Species identification with confidence level (when possible)
4. Brief biosecurity risk assessment

METADATA HANDLING:
Base analysis solely on visual image content. Do not use EXIF data, timestamps, or embedded metadata.

User Prompt:

Analyze this marine biofouling image using the standardized Level of Fouling (LoF) classification system. Base your analysis solely on visual content.

ASSESSMENT INSTRUCTIONS:
Apply the official LoF decision tree methodology:
1. Examine the visible surface for any fouling
2. Determine if fouling is: absent, slime-only, or includes macrofouling
3. If macrofouling present, estimate percent cover of visible surface
4. Apply appropriate LoF rank:
   - LoF 0: No fouling visible
   - LoF 1: Slime layer only (any amount), no macrofouling
   - LoF 2: 1-5% macrofouling coverage
   - LoF 3: 6-15% macrofouling coverage  
   - LoF 4: 16-40% macrofouling coverage
   - LoF 5: 41-100% macrofouling coverage

COVERAGE ESTIMATION:
- Focus on actual biological organisms covering the surface
- Consider the entire visible area when calculating percentages
- Distinguish biological growth from discoloration, shadows, or surface marks

Provide your analysis in this exact format:
**LoF Rating:** [0-5] - [description]
**Coverage:** [percentage] of visible surface
**Species:** [identification with confidence level]
**Risk:** [biosecurity assessment]
\end{Verbatim}
\end{tcolorbox}

\vspace{1em} 

\begin{tcolorbox}[colback=gray!5,colframe=black!50,
                  title=Baseline Prompt (First LLM Test),
                  breakable,width=\textwidth]
\label{app:baseline-prompt}
\begin{Verbatim}[breaklines=true]
You are a marine biofouling expert applying the Level of Fouling (LoF) rank scale developed by Davidson et al. (2019).

Classify the following image using the LoF scale, which ranges from 0 to 5 and is based on **visible percent cover of macrofouling** on submerged vessel surfaces (e.g., hulls, rudders, propellers):

LoF 0: No fouling at all — clean surface with no slime or macrofouling
LoF 1: Only a slime layer (biofilm), no visible macrofouling
LoF 2: 1–5% of surface covered with isolated macrofouling patches
LoF 3: 6–15% of surface covered with moderate macrofouling
LoF 4: 16–40% of surface covered with extensive macrofouling
LoF 5: 41–100% of surface covered with very heavy macrofouling

Definitions:
- **Slime layer** is a microscopic film, not visible as organisms
- **Macrofouling** includes visible organisms like barnacles, seaweed, sponges, etc.
- Percent cover refers to the *visible surface* in the image, not the whole vessel

Please return:
1. A **LoF rank (0–5)** for the image
2. A short **justification** based on percent cover and fouling type
3. A note on any visible **invasive species**, if present

Do not say “unable to classify.” Always make a classification based on what is visible, even if uncertain. Assume the image is a valid vessel surface. Focus on the biofouling cover, not species identity.
\end{Verbatim}
\end{tcolorbox}

\twocolumn

\end{document}